\documentclass[10pt,twocolumn,letterpaper]{article}

\usepackage{cvpr}
\usepackage{times}
\usepackage{epsfig}
\usepackage{graphicx}
\usepackage{amsmath}
\usepackage{amssymb}
\usepackage{subfigure}
\usepackage{overpic}
\usepackage{algorithm, algorithmic}

\usepackage{enumitem}
\setenumerate[1]{itemsep=0pt,partopsep=0pt,parsep=\parskip,topsep=5pt}
\setitemize[1]{itemsep=0pt,partopsep=0pt,parsep=\parskip,topsep=5pt}
\setdescription{itemsep=0pt,partopsep=0pt,parsep=\parskip,topsep=5pt}

\usepackage[pagebackref=true,breaklinks=true,letterpaper=true,colorlinks,bookmarks=false]{hyperref}

\cvprfinalcopy 

\graphicspath{{figures/}}

\begin{document}

\newcommand{\figref}[1]{图\ref{#1}}
\newcommand{\tabref}[1]{表\ref{#1}}
\newcommand{\equref}[1]{式\ref{#1}}
\newcommand{\secref}[1]{第\ref{#1}节}

\title{Iterative Global Similarity Points : A robust coarse-to-fine integration solution for pairwise 3D point cloud registration }

\author{Yue Pan, Bisheng Yang, Fuxun Liang and Zhen Dong\\
Wuhan University \\
Wuhan, China, 430079\\
{\tt\small \{panyue, bshyang, liangfuxun, dongzhenwhu\}@whu.edu.cn}
}

\maketitle

\begin{abstract}
   In this paper, we propose a  coarse-to-fine integration solution inspired by the classical ICP algorithm, to pairwise 3D point cloud registration with two improvements of hybrid metric spaces (\eg, BSC feature and Euclidean  geometry spaces) and globally optimal correspondences matching. First, we detect the keypoints of point clouds and use the Binary Shape Context (BSC)\cite{67} descriptor to  encode their local  features. Then, we formulate the correspondence matching task as an energy function, which models the global similarity of keypoints  on the hybrid spaces of BSC feature and Euclidean  geometry. Next,  we estimate the globally optimal correspondences through optimizing the energy function by  the Kuhn-Munkres\cite{68} algorithm and then calculate the transformation based on the correspondences. Finally, we  iteratively refine the transformation between two point clouds by conducting optimal correspondences matching and transformation calculation  in a mutually reinforcing manner, to achieve the coarse-to-fine registration under an unified framework. The proposed method is evaluated and compared to several state-of-the-art methods on selected  challenging datasets with repetitive, symmetric and incomplete structures. Comprehensive experiments demonstrate that the proposed IGSP algorithm obtains good performance and outperforms  the state-of-the-art methods in terms of both rotation and translation  errors.
\end{abstract}

\section{Introduction}

As the development of laser scanning and photogrammetry, point cloud, which depicts the world in 3D manner, has been widely collected by many platforms like airborne laser scanning (ALS) and terrestrial laser scanning (TLS). Point cloud registration is a fundamental problem in 3D computer vision and photogrammetry. Given several sets of points in different coordinate systems, the aim of registration is to find the  transformation that best aligns all of them into a common coordinate system. Point cloud registration plays a significant role in many vision applications such as 3D model reconstruction \cite{1}\cite{4}, cultural heritage management\cite{5}\cite{6}, landslide monitoring\cite{8} and solar energy analysis\cite{9}.In 3D object recognition, fitness degree between an existing model object and an extracted object in the scene can be evaluated with registration results \cite{13}\cite{14}.In the field of robotics,  for simultaneous localization and mapping(SLAM), the  registration can act as a visual odometry to realize structure from motion and locate the current view into the global scene\cite{18}\cite{17}. 

Generally, there are four key challenges for unordered point clouds registration: (1) uneven point densities , (2) the huge amount of data , (3) repetitive, symmetric, and incomplete structures , and (4) limited overlaps between point clouds . All of these challenges can seriously affect the performance of point cloud registration methods. To address these challenges, extensive studies have been done to improve the accuracy, efficiency, and robustness of point cloud registration. Point cloud registration can be roughly categorized into pairwise and multi-view registration according to the number of input point clouds\cite{19}. The pairwise registration is the prerequisite of multi-view registration, which is the focus of this paper.  The registration process can be further divided into two major steps: coarse registration, in  which an initial transformation between two point clouds is estimated, and fine registration, in which the initial transformation is then further refined\cite{20}. 

The remainder of this paper is organized as follows. Following this introduction, Section 2 briefly reviews the representative work related to pairwise point cloud registration and introduces the paper's contribution.  Section 3 gives a detailed description of the proposed pairwise point cloud registration method. The proposed method is validated in experimental studies in Section 4. Finally, the conclusions and future research directions are presented in Section 5.

\section{Related work}
\subsection{Coarse Registration.}
Normally, a pairwise coarse registration method has the following procedures. First,  key primitive elements (\eg, points, lines, and planes) are detected from each point cloud, where the point-based methods are more popular due to their feasibility to different scenes\cite{28}. The keypoint detectors (\eg, local surface patches\cite{31}, 2.5D SIFT\cite{33}, 3D SURF\cite{34} and 3D Harris\cite{35}) are exploited to extract keypoints from raw point clouds.
Second, the  feature descriptors (\eg, Spin image\cite{36}, Fast Point Feature Histograms (FPFH)\cite{38}, Viewpoint Feature Histogram (VFH)\cite{39} and Rotational Projection Statistics (RoPS)\cite{41}) are calculated to encode the  local shape information of each keypoint. Recently, \cite{43} applies a deep neural network auto-encoder to realize the same effect.
Third, various feature matching strategies are applied to determine the initial correspondences (\eg, reciprocal correspondence\cite{44}, correlation coefficient\cite{45}, and chi-square test\cite{46}). Finally, due to the fact that some of the obtained correspondences are incorrect, outliers need to be filtered from the initial correspondences, calling for a method to eliminate outliers  and calculate the transformation between point clouds based on the remaining correspondences. This can be solved by some robust transformation estimation algorithms (\eg, Random Sample Consensus (RANSAC)\cite{47} , 3D Hough voting\cite{48}, geometric consistency constraints based on line distance or triangle property\cite{50}\cite{25}, and Game Theory based matching algorithms \cite{52}\cite{54}.)

Additionally, there are some point-based methods which do not follow the abovementioned workflow. For example, the 4-Points Congruent Sets (4PCS)\cite{4PCS} and its variants (\eg, SUPER-4PCS\cite{Super4PCS}, Keypoint-based 4PCS (K-4PCS)\cite{K4PCS})determines the corresponding four-point base sets by taking intersection ratios of these four points instead of using feature descriptor for matching, thus improving the efficiency of RANSAC based global registration to a great extent.

Though most of the  coarse registration algorithms can generally provide satisfactory registration results, they still have limitations. For symmetric and large-scale point clouds,  a robust and efficient correspondence matching algorithm, which is capable of acquiring the globally optimal correspondences,  is urgently needed.

\subsection{Fine Registration.}
As for fine registration, the Iterative Closest Point (ICP) algorithm \cite{62} and its variants\cite{63}\cite{64} are the most commonly used  methods, which alternate between correspondence matching and transformation calculation until convergence. \cite{63} proposed a Geometric Primitive ICP with Random sample consensus (GPICPR) in which the local surface normal vector and geometric curvature are used for matching and neighborhood searching.\cite{64} used geometric features to improve the classical ICP algorithm. ICP and these variants are able to acquire registration results with high efficiency and accuracy. However, they require a good initialization to avoid  converging to bad local minimum. 

\subsection{Contribution}
To overcome the limitations and challenges, this paper proposes an Iterative Global Similarity Points (IGSP) algorithm to  realize a coarse-to-fine integration solution to pairwise 3D point cloud registration. IGSP is inspired by the classical ICP algorithm with two improvements of hybrid metric spaces (\eg, BSC feature and Euclidean  geometry spaces) and globally optimal correspondences matching. 

Specifically, the main contributions of the proposed method are as follows:
i) we formulate the correspondence matching task as an energy function, which models the  global similarity of keypoints on the hybrid metric spaces of BSC feature and Euclidean  geometry, to get a more robust result .
ii).we realize a coarse-to-fine registration by conducting  optimal correspondences matching and transformation calculation  in an iterative and mutually reinforcing manner, so that a good initialization is not essential.
\section{Method}


\begin{figure*}
\begin{center}
\includegraphics[width=1.0\linewidth]{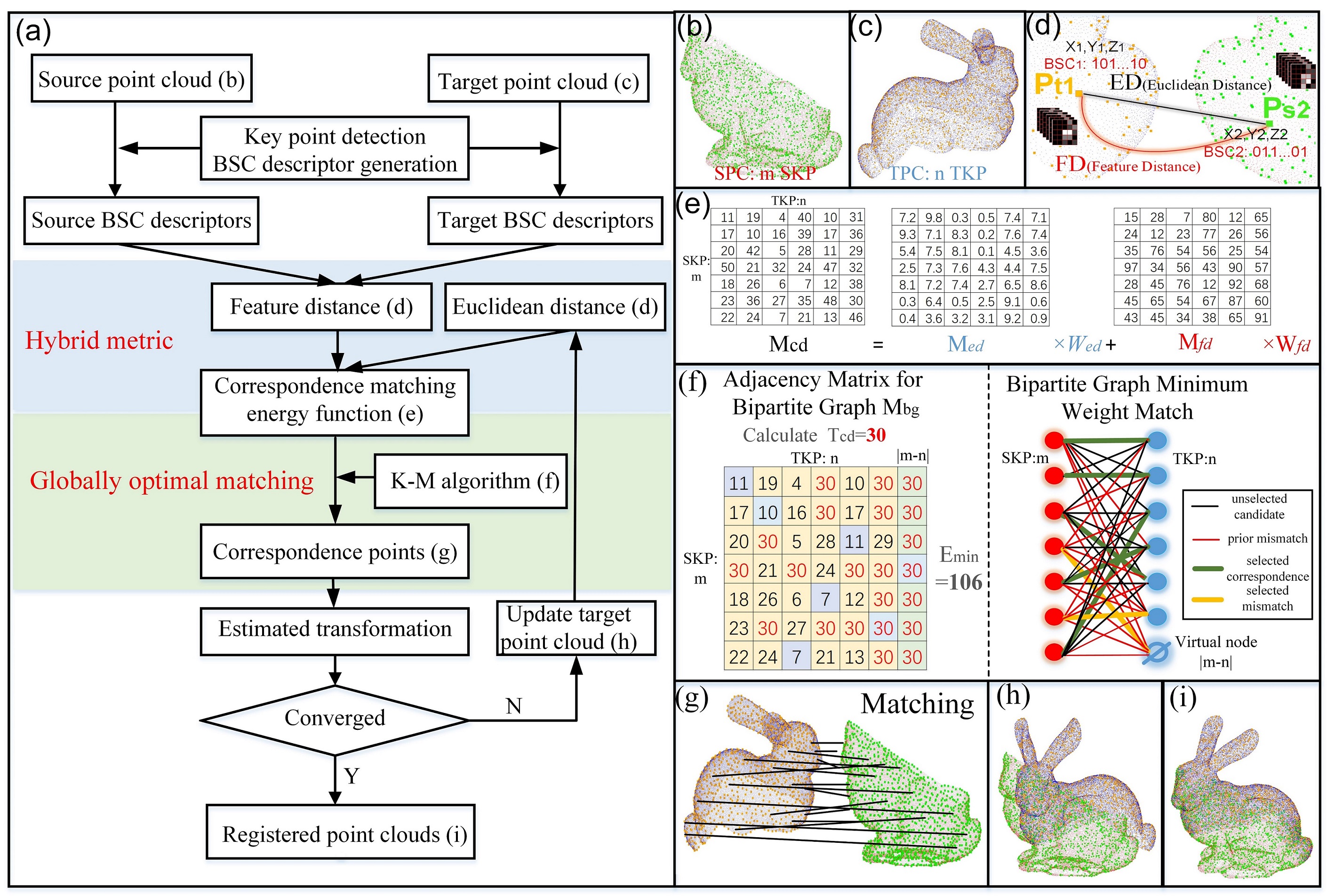}
\end{center}
   \caption{Overview of the proposed IGSP algorithm framework:(a)workflow, (b) the input Source Point Cloud (SPC, in red) and detected  Source Keypoints (SKP, in green), (c) the input Target Point Cloud (TPC, in purple) and detected Target Keypoints (TKP, in orange), (d) a sketch map for the Euclidean Distance (ED) and Feature Distance (FD) (e) the pairwise Compound Distance (CD) and the corresponding ED matrix, FD matrix, (f) Left:an example of energy function optimization by KM algorithm. Numbers in grids represent the compound distance between  keypoints. The mismatch threshold $T_{cd}$ (in red) is firstly calculated (30), and a virtual column (the rightmost column) is inserted. Red numbers are the prior mismatches. The numbers  in blue grids are used to calculate the result (106), which is considered as the minimum matching energy $E_{\min}$. Right:the corresponding sketch map of the example. SKP and TKP are shown in red  and blue respectively. The virtual node is represented by $\varPhi $. Four different type of lines are illustrated (black: unselected candidate, red: prior mismatch, green: selected correspondence, orange: selected mismatch) (g) the correspondences solved by KM algorithm, (h) the updated result after several iterations, (i) final registered result.}
\label{fig:1}
\end{figure*}

\subsection{IGSP framework}
An overview of the proposed IGSP algorithm is shown in Fig.\ref{fig:1}. Similar to ICP, an iteration process is also involved. IGSP iteratively conducts  globally optimal correspondences matching and transformation calculation, until the transformation is negligible. The  workflow of  IGSP  is shown in Fig.\ref{fig:1}a. First, we detect the keypoints of point clouds and generate the Binary Shape Context (BSC)\cite{67} descriptors to  encode their local  features. Then, we formulate the correspondence matching task as an energy function, which models the global similarity of keypoints on the hybrid metric space of BSC feature and Euclidean geometry, as shown in Fig.\ref{fig:1}d-e. Next,  we determine the globally optimal correspondences by optimizing the energy function by  the Kuhn-Munkres(KM) algorithm\cite{68} and then calculate the transformation based on them, as shown in Fig.\ref{fig:1}f-g. Finally, we refine the transformation between two point clouds by iteratively conducting optimal correspondences matching and transformation calculation  to realize a coarse-to-fine registration under an unified framework, as shown in Fig.\ref{fig:1}h-i. The pseudo code of  IGSP  is  shown in Algorithm \ref{alg:1}.

\begin{algorithm}
	
    \renewcommand{\algorithmicrequire}{\textbf{Input:}}
	\renewcommand{\algorithmicensure}{\textbf{Output:}}

	\caption{Iterative Global Similarity Points (IGSP)}
	\label{alg:1}
    {\bf Notation:}\\
    $SPC,TPC:$Source and Target Point Clouds\\
    $SKP\left( sk_1,sk_2,...sk_m \right),TKP\left( tk_1,tk_2,...tk_n \right):$Source and Target Keypoint sets \\
    $SBSC\left( sb_1,sb_2,...sb_m \right),TBSC\left( tb_1,tb_2,...tb_n \right):$BSC features extracted from $SKP$ and $TKP$ \\
    $M_{ed},M_{fd},M_{cd},M_{bg}:$Matrix of ED, FD, CD and the bipartite graph (ED, FD and CD represents Euclidean, feature and compound distance respectively)\\
    $\sigma _r,\sigma _t:$convergence threshold for translation and rotation variation\\
    $M:$the set of correspondences in $SKP$ and $TKP$\\
    $Rt_{temp}:$ transformation estimated at the current iteration \\
   $Rt:$ transformation accumulated from the first to the current iteration\\
  $k:$iteration number 

	\begin{algorithmic}[1]
        \REQUIRE $SPC$ and $TPC$
		\ENSURE  $Rt$
		\STATE Detect $m$ keypoints $SKP\left( sk_1,sk_2,...sk_m \right)$ from $SPC$ and $n$ keypoints $TKP\left( tk_1,tk_2,...tk_n \right)$ from $TPC$
		\STATE Extract BSC feature $SBSC\left( sb_1,sb_2,...sb_m \right)$ of $SKP\left( sk_1,sk_2,...sk_m \right)$ and $TBSC\left( tb_1,tb_2,...tb_n \right)$ of  $TKP\left( tk_1,tk_2,...tk_n \right)$
		\STATE Calculate the $m\times n$ matrix $M_{fd}$ using $SBSC\left( sb_1,sb_2,...sb_m \right)$ and $TBSC\left( tb_1,tb_2,...tb_n \right)$, as Eq.\ref{eq:4}
        \STATE Initialization:$k\gets \text{0,}Rt\gets I,\varDelta r\gets \infty ,\varDelta t\gets \infty$
		\WHILE{$\varDelta r>\sigma _r\land \varDelta t>\sigma _t$}
		\STATE Calculate the $m\times n$ matrix $M_{ed}$ using $SKP\left( sk_1,sk_2,...sk_m \right)$ and $TKP\left( tk_1,tk_2,...tk_n \right)$, as Eq.\ref{eq:3}
		\STATE Calculate $W_{ed}$ and $W_{fd}$, as Eq.\ref{eq:7}
		\STATE $M_{cd}\gets W_{ed}M_{ed}+W_{fd}M_{fd}$
		\STATE Calculate $T_{cd}$, as Eq.\ref{eq:11}
		\STATE Calculate $\max \left( m,n \right) \times \max \left( m,n \right)$ matrix $M_{bg}$ using $M_{cd}$ and $T_{cd}$, as Eq.\ref{eq:5}
		\STATE $M=KM\left( M_{bg},T_{cd} \right)$, $KM$ represents Kuhn–Munkres algorithm, which takes the bipartite graph adjacency matrix and the mismatch threshold as input and outputs the matching set.
        \STATE Calculate $Rt_{temp}$ from $M$ using $SVD$, as Eq.\ref{eq:9}
        \STATE $TKP\gets Rt_{temp}TKP$
        \STATE Calculate $\varDelta r$ and $\varDelta t$ from $Rt_{temp}$
        \STATE $Rt\gets Rt_{temp}Rt,\ k\gets k+1$
	    \ENDWHILE
	\end{algorithmic}
    \label{alg:1}
\end{algorithm}

\subsection{Energy function construction}
In this section, a global energy function for keypoints matching is constructed, as Eq.\ref{eq:1}, which consists of data cost and penalty cost. Data cost represents the global similarity of matched keypoints in source and target point clouds, while penalty cost indicates the number of keypoints without correspondence. For data cost, $p$ and $q$ represent a matching keypoint pair. $M$ is the  set of correspondences between two point clouds and $CD(p,q)$ means the compound distance between keypoint $p$ and $q$ ,which is defined as a weighted sum of  feature distance $FD(p,q)$ and Euclidean distance $ED(p,q)$ , as shown in Eq.\ref{eq:2}. $W_{ed}$ and $W_{fd}$ are the weights of  $ED$ and $FD$. As shown in Eq.\ref{eq:4}, $FD(p,q)$ is defined as the Hamming distance between the BSC descriptors of $p$ and $q$ .  $ED(p,q)$ is calculated as a scale factor related to point density $s_{ed}$ times the Euclidean distance between $p$ and $q$ , as shown in Eq.\ref{eq:3}. $W_{ed}$ and $W_{fd}$  are assigned as Eq.\ref{eq:7}, in which $k$ is the iteration number counted from zero and $m$ controls the weight changing rate.  For penalty cost in Eq.\ref{eq:1}, $\varphi $ is the set of unmatched keypoints in two point clouds and $\left| \varphi \right|$ is the  number . $W_p$ is the weight of penalty cost, which will be further used as the criterion for mismatch judgement in 3.3. 
\begin{equation}
E=\underset{Data\_cost}{\underbrace{\sum_{p\in S,q\in T,\left( p,q \right) \in M}{CD\left( p,q \right)}}}+\underset{Penalty\_cost}{\underbrace{W_p\left| \varphi \right|}}
 \label{eq:1}
\end{equation}

\begin{equation}
CD\left( p,q \right) = W_{fd}FD\left( p,q \right)+W_{ed}ED\left( p,q \right)
\label{eq:2}
\end{equation}

\begin{equation}
FD\left( p,q \right) =HD\left( f_p,f_q \right)
\label{eq:4}
\end{equation}
\begin{equation}
ED\left( p,q \right) =s_{ed}\lVert p-q \rVert
\label{eq:3}
\end{equation}\begin{equation}
\begin{cases}
	W_{fd}=e^{-\frac{k}{m}}\\
	W_{ed}=1-e^{-\frac{k}{m}}\\
\end{cases}\ k=\text{0,1,}2...
\label{eq:7}
\end{equation}

By minimizing the energy function, we can obtain the globally optimal correspondences $\left\{ M,\varphi \right\}^{*}$, as shown in Eq.\ref{eq:8}.
\begin{equation}
\begin{split}
&\left\{ M,\varphi \right\}^{*}=\underset{\left\{ M,\varphi \right\}}{arg\min}E=\underset{penalty\_cost}{\underbrace{W_p\left| \varphi \right|}}+\\
&\underset{Data\_cost}{\underbrace{\sum_{p\in S,q\in T,\left\{ p,q \right\} \in M}{\left( \left( 1-e^{-\frac{k}{m}} \right) ED\left( p,q \right) +e^{-\frac{k}{m}}FD\left( p,q \right) \right)}}}
\end{split}
\label{eq:8}
\end{equation}

\subsection{Energy function optimization by KM algorithm}
In this section, the global energy function Eq.\ref{eq:8}  is generalized into a bipartite graph minimum weight match problem and then solved by using slacked KM algorithm.

A bipartite graph is a graph whose vertices can be divided into two disjoint sets $S$ and $T$\cite{74}. Each edge connects a vertex in $S$ to a vertex in $T$ in this graph. Given a bipartite graph and its corresponding edges, the optimal weight matching guarantees that each node in one sub-graph can be matched to only one node in the other subgraph, and this matching can achieve global matches with the maximal or minimum summation of  edge weight.

The task to find the best matching for two keypoint sets can be modeled as an optimal matching task of bipartite graph. In the weighted bipartite graph $G=\left( S,T,E \right)$, each keypoint in source and target cloud is represented by one node respectively in set $S$ and $T$. Suppose there are $m$  and $n$ keypoints detected from source and target cloud respectively, and when $m\ne n$  , we add $\left| m-n \right|$ virtual unmatched nodes $N_v$ to the set with less nodes to make $\left| S \right|=\left| T \right|=\max \left( m,n \right)$. Each edge $e\left( p,q \right) \in E$ corresponds to a distance between node $p$ in $S$ and node $q$ in $T$. With $T_{cd}$ as the mismatch threshold for $CD$, $e(p,q)$ is computed as follows:
\begin{equation}
e\left( p,q \right) =\begin{cases}
	CD\left( p,q \right) ,CD\left( p,q \right) <T_{cd}\land p\notin N_v\land q\notin N_v\\
	T_{cd}\ \ \ \ \ \ ,else\\
\end{cases}
\label{eq:5}
\end{equation}

  Next, the sum of all edge weights can be minimized as $E_{\min}^{bgm}$ in  Eq.\ref{eq:8}. When  $T_{cd}=2W_p$  ,  there is only a constant difference between  $E_{\min}^{bgm}$ and the minimum energy $E_{min}$ for the energy function, so that the minimum weight match is equivalent to the optimization of  energy function.  The selected edges whose weight $e^*\left( p,q \right) <T_{cd}$ make up the optimal matching set $M^*$, and the unmatched keypoints set $\varphi ^*$ is also determined.
\begin{equation}
\begin{split}
&E_{\min}^{bgm}=\sum{e^*\left( p,q \right)}+T_{cd}\left( \max \left( m,n \right) -\left| M^* \right| \right) 
\\
&=\sum{e^*\left( p,q \right)}+W_p\left( m+n+\left| m-n \right|-2\left| M^* \right| \right) 
\\
&=\sum_{p\in S,q\in T,\left\{ p,q \right\} \in M^*}{CD\left( p,q \right)}+W_p\left| \varphi \right|+W_p\left| m-n \right|
\\
&=E_{\min}+W_p\left| m-n \right|
\end{split}
\label{eq:6}
\end{equation}
      Given such weighted bipartite graph $G=\left( S,T,E \right)$, the Kuhn-Munkres(KM) algorithm is employed. It  outputs a complete bipartite matching  with minimum matching weight by transforming the problem from an optimization problem of finding a minimum weight matching into a combinatorial one of finding a perfect matching. For efficiency concern, we apply the KM algorithm with slacked terms whose time complexity is $O\left( \left| V \right|^3 \right)$, in which $\left| V \right|$ is the number of vertexes of the graph. Fig.\ref{fig:1} illustrates a simple example of this process.

It is proved that given a point cloud pair with $m$ and $n$ keypoints respectively and a fixed threshold $W_p$ or $T_{cd}$ , $E_{min}$ can be calculated by solving $E_{\min}^{bgm}$ via KM algorithm.

Once the correspondence is determined, the optimal transformation can be estimated via Singular Value Decomposition(SVD), as shown in Eq.\ref{eq:9} .
\begin{equation}
\left\{ R|t \right\}^* =\underset{\left\{ R,t \right\}}{arg\min}J=\sum_{p\in S,q\in T,\left\{ p,q \right\} \in M^*}{\lVert p-Rq+t \rVert ^2}
\label{eq:9}
\end{equation}
\subsection{Iteration Process}
Although a correspondence set $\left\{ M,\varphi \right\}$ with global similarity can be solved efficiently,  it is not robust enough. Inspired by ICP, we propose an iteration process by alternating between solving for the correspondence using the method presented above and estimating the transformation using SVD until convergence. For each iteration, $W_{ed}$ and $W_{fd}$ are updated according to Eq.\ref{eq:7}. In the iteration process, $W_{fd}$ decreases from 1 to 0, while $W_{ed}$ increases from 0 to 1.  At first, $W_{ed}=0$  and the correspondence is estimated only based on keypoints’ feature descriptor, thus producing a coarse registration result. As the process continues,  the weight of Euclidean distance increases,  introducing the geometric restriction and refining the registration over and over.


As mentioned before,  $W_p$  (or $T_{cd}$) is the weight of penalty cost used as the criterion for mismatch judgement. In the context of bipartite graph minimum weight match, candidate pairs whose $CD>T_{cd}$ would be regarded as mismatches. A self-adaptive scheme for the determination of this subtle parameter is devised, as shown in Eq.\ref{eq:11}.

\begin{equation}
T_{cd}=\begin{cases}
	\mu _{cd}+p_{1}^{t}\sigma _{cd}&		,k=0\\
	p_{2}^{t}\left( 1-e^{-\frac{k}{m}} \right) \overline{ED_{k-1}^{c}}+p_{3}^{t}e^{-\frac{k}{m}}\overline{FD_{k-1}^{c}}&		,k>0\\
\end{cases}
\label{eq:11}
\end{equation}

When $k=0$, given a  pairwise match matrix of $CD$ , the mean value $\mu _{cd}$ and standard deviation $\sigma _{cd}$ can be calculated. With a threshold parameter $p_1^t$ , we get the initial $T_{cd}$. When $k>0$,  $\overline{ED_{k-1}^{c}}$ and $\overline{FD_{k-1}^{c}}$ are the average Euclidean and feature distance of matched keypoints pairs of last iteration. Besides, $p_2^t$ and $p_3^t$  are two threshold parameter to be determined. The search space of candidate match keeps narrowing under the restriction of the average geometric and feature similarity of last iteration’s correspondence set. Generally, the larger parameters $p_{i}^{t}\left( i=\text{1,2,}3 \right)$ are, the larger the search space of candidate match is.

Finally, as for the condition of convergence, we set thresholds for both translation and rotation in practice. Once the transformation difference between two iterations meets the condition, the iterative process stops. 

\section{Experiment}
\subsection{Experimental setup}
{ \bf Experiment Platform.}
 The experiments are implemented  with a 16 GB RAM and an Intel Core i7-6700HQ @ 2.60GHz CPU. The proposed  IGSP algorithm and the compared baselines are all implemented in C++ with the help of point cloud library (PCL) \cite{80}.

\begin{figure}[t]
\begin{center}
\includegraphics[width=1.0\linewidth]{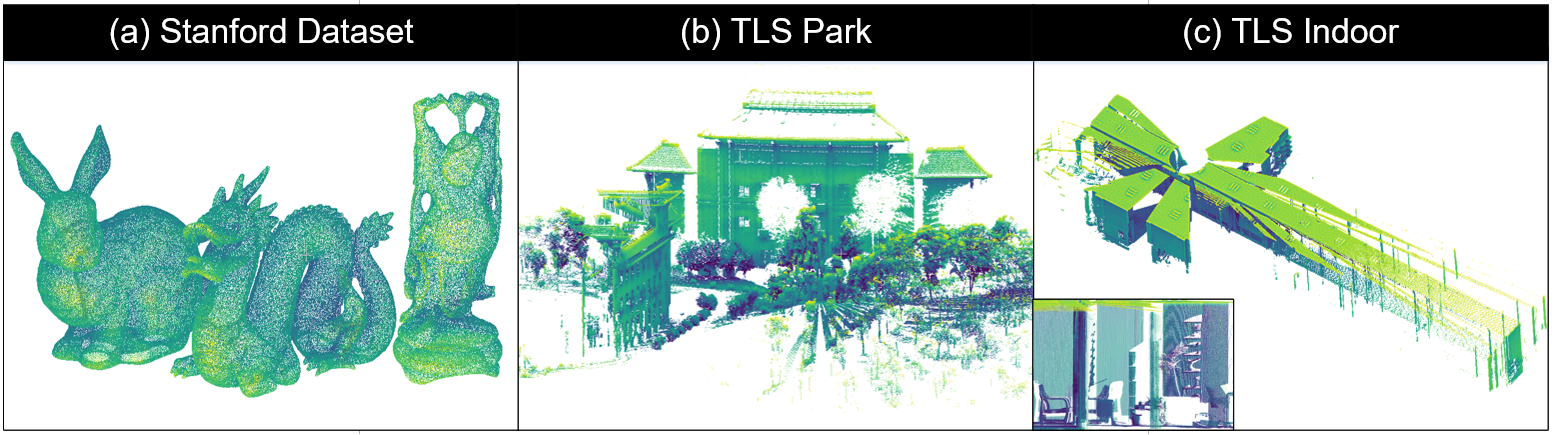}
\end{center}
\caption{The sampled point clouds from three datasets, rendered by ambient occlusion. (for indoor datasets, some details are also shown)}
\label{fig:2}
\end{figure}

{ \bf Datasets description.}
As shown in Fig.\ref{fig:2}, the performance of the proposed IGSP algorithm is evaluated both on small scale scans of 3D models (Stanford Repository\cite{76}) and on challenging real-world TLS datasets (\eg Park, Indoor). These TLS datasets are challenging as (i) each of the datasets contains several scans and billions of points (ii) repetitive, symmetric and incomplete structures are common.

\begin{figure*}[htbp]
\begin{center}
\includegraphics[width=1.0\linewidth]{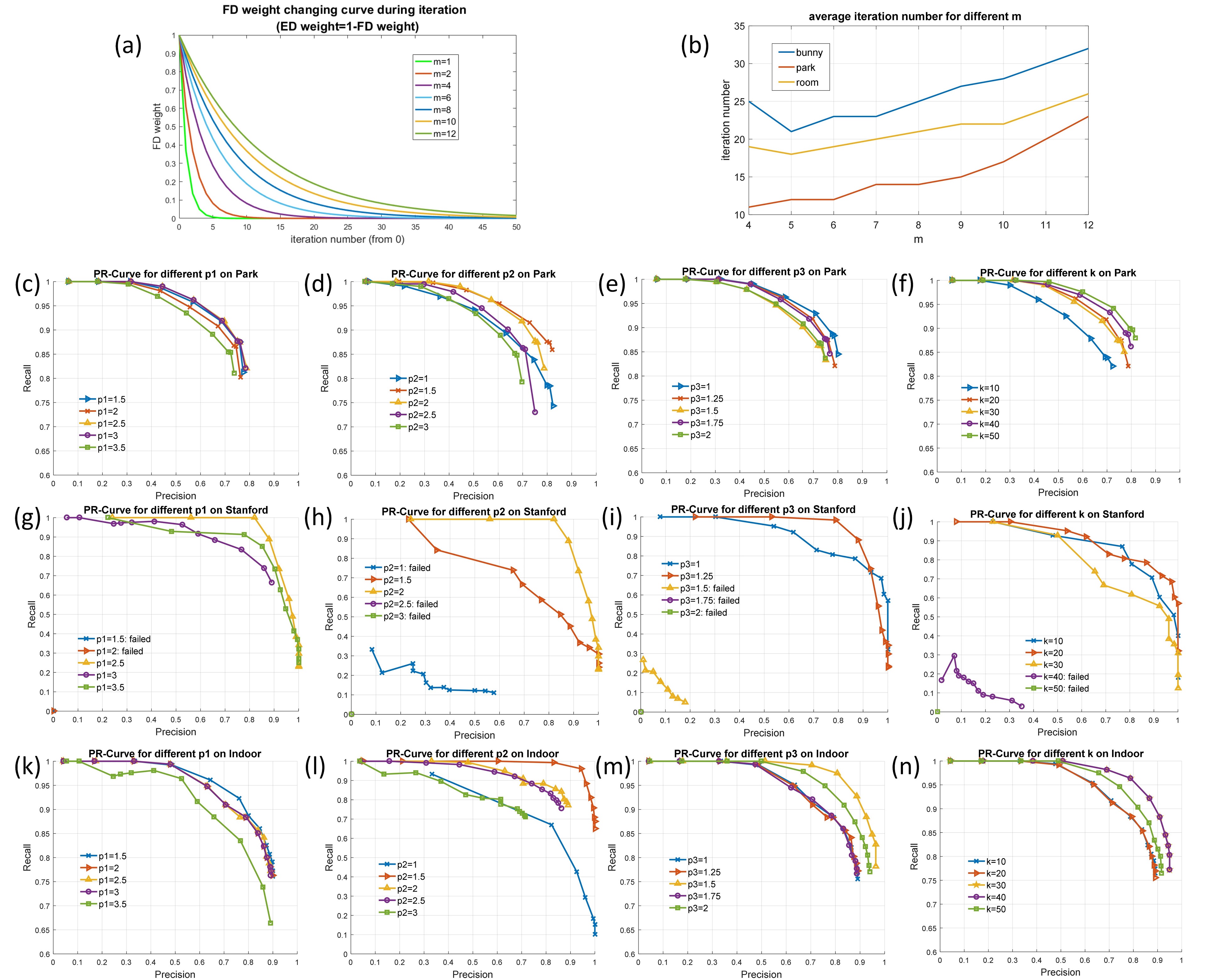}
\end{center}
   \caption{Parameter settings experiment result: (a-b) m's effect on ED, FD's weight and total iteration number for IGSP. (c-n) precision and recall performance of different parameter settings for IGSP's correspondence grouping on Park, Standford and Indoor datasets respectively.}
\label{fig:3}
\end{figure*}

{ \bf Evaluation criteria.}
We evaluate the performance of the proposed IGSP method in terms of rotation error and translation error which are commonly used for the evaluation of point cloud registration\cite{41}.
Given the estimated transform $T$ and ground truth transform $T^G$ , rotation error $e^r$ and translation error $e^t$ can be calculated as Eq.\ref{eq:12},\ref{eq:13}.

\begin{equation}
 \varDelta T=T\left( T^G \right) ^{-1}=\left[ \begin{matrix}
	\varDelta R&		\varDelta t\\
	0&		1\\
\end{matrix} \right]
\label{eq:12}
\end{equation}

\begin{equation}
\begin{cases}
	e^r=\arccos \left( \frac{tr\left( \varDelta R \right) -1}{2} \right)\\
	e^t=\lVert \varDelta t \rVert\\
\end{cases}
\label{eq:13}
\end{equation}

Besides, the keypoint correspondences’ quality can be evaluated with precision and recall as Eq.\ref{eq:14}, where $TP$ is the number of true positive correspondences, $FP$ is the number of false positive ones, and $FN$ is the number of false negative ones.

\begin{equation}
\begin{cases}
	precision=\frac{TP}{TP+FP}\\
	recall=\frac{TP}{TP+FN}\\
\end{cases}
\label{eq:14}
\end{equation}

{ \bf Parameter settings.}
To get the result of correspondence grouping step with relatively high recall and precision, the parameters should be set reasonably. As efficiency is another important concern, parameters that result in less total iteration number would be preferred. Experiment results on three datasets are shown in Fig.\ref{fig:3}, from which IGSP's main parameters are determined in consideration of the aforementioned criteria. Table \ref{tab:1} shows the parameter settings of the proposed IGSP method. As for the denotation, $r_k$ is the Non-maximum suppression radius of BSC. The parameter settings are used for all the experiments in this paper.

\begin{table}
\caption{Parameter Settings of the proposed IGSP method.}
\begin{center}
\begin{tabular}{lll}
\hline
Parameter&Description&Value\\
\hline
$s_{ed}$&Scale factor of ED& $\frac{k}{r_k},k=30$\\
\hline
m&Iterative weight changing rate&8\\
\hline
$p_{1}^{t}$&Initial threshold parameter&2.5\\
\hline
$p_{2}^{t}$&Threshold parameter for ED&1.5\\
\hline
$p_{3}^{t}$&Threshold parameter for FD&1.25\\
\hline
\end{tabular}
\end{center}

\label{tab:1}
\end{table}

\subsection{Results, evaluation and analysis}

\begin{figure*}[htbp]
\begin{center}
\includegraphics[width=1.0\linewidth]{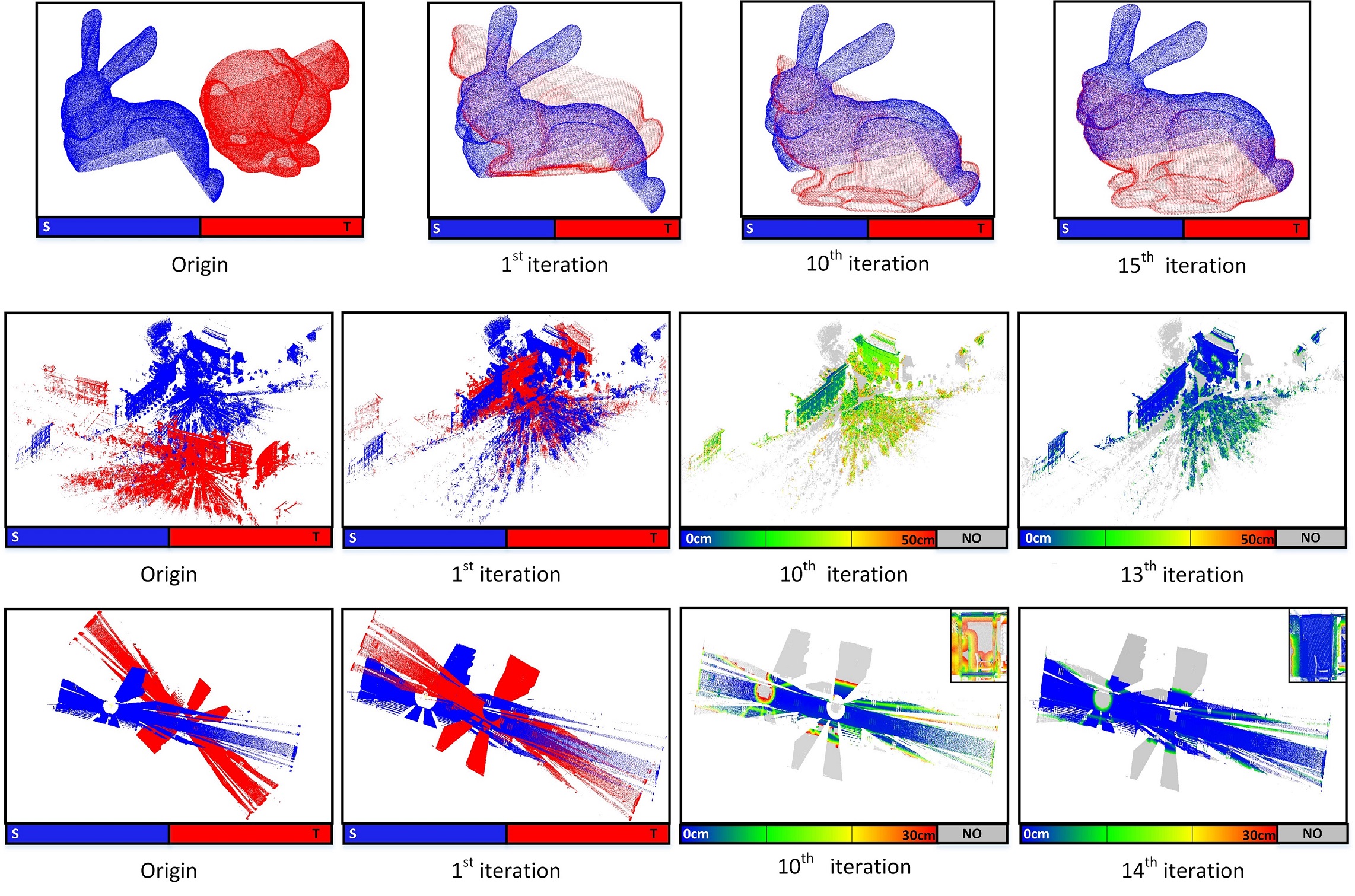}
\end{center}
     \caption{Registration result of (a)Stanford Bunny, (b)TLS Park and (c)TLS Indoor datasets during the iterative process. Source and target point clouds are shown in blue and red respectively. The Gradient color indicate the cloud to cloud distance of overlapping area and the non-overlapping points are shown in gray. }
\label{fig:4}
\end{figure*}
\begin{figure}[htb]
\begin{center}
\includegraphics[width=1.0\linewidth]{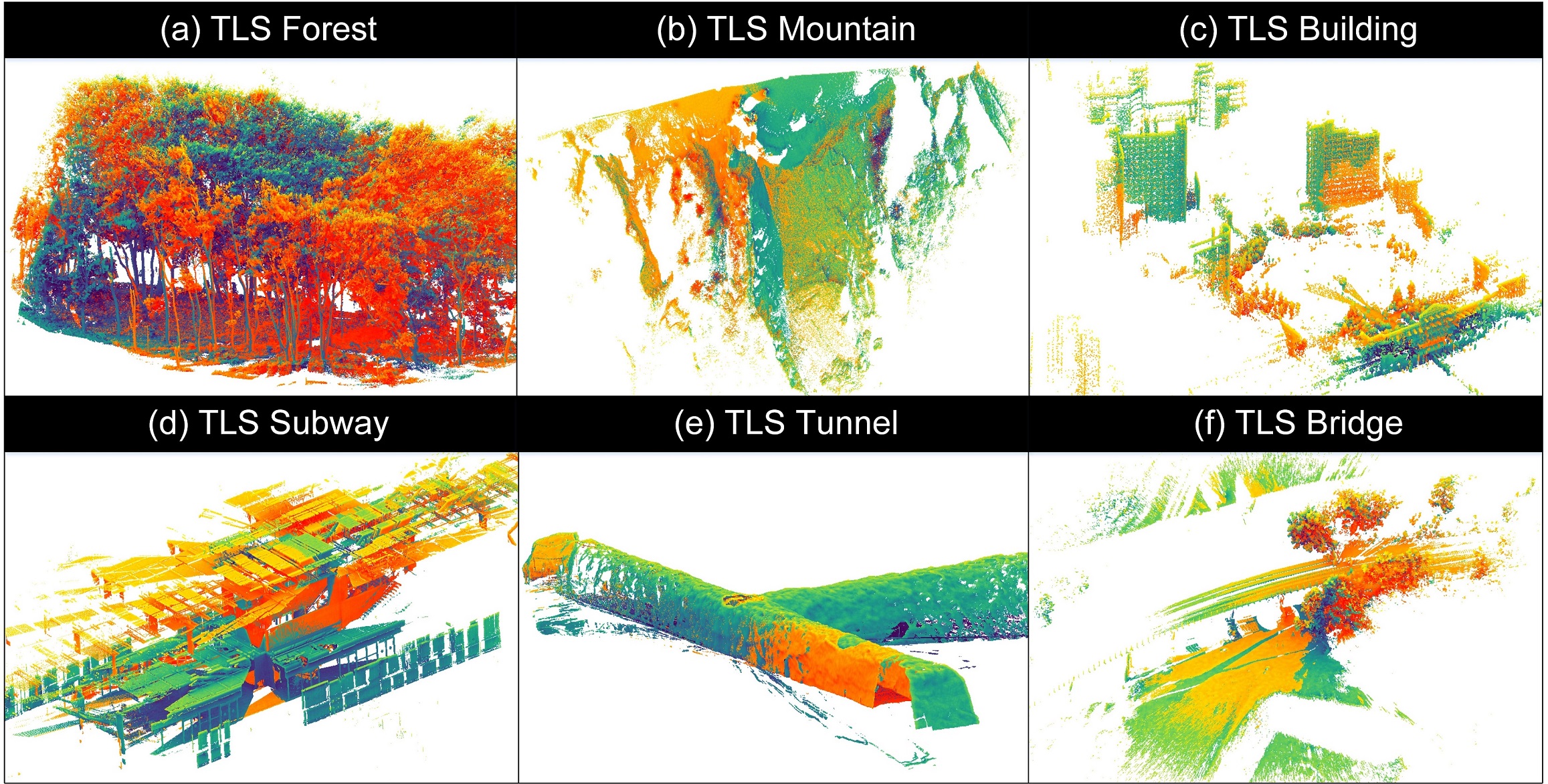}
\end{center}
   \caption{Registration result of other challenging real-world datasets}
\label{fig:9}
\end{figure}

{ \bf Registration results.}
Fig.\ref{fig:4} show different phases of registration results using the proposed IGSP method on three testing datasets respectively. As seen in these figures,by the IGSP method, the point cloud pairs are iteratively converged from coarse to fine and get registered successfully. To further test the robustness, the algorithm is applied on other challenging real-world datasets, as shown in Fig.\ref{fig:9}. These results show that the proposed IGSP method performs well for various scenes, including those with repetitive, symmetric, noisy and incomplete structures, which are quite challenging for previous methods.

{ \bf Registration accuracy evaluation.}
Average rotation error and translation error on three datasets are reported in Table \ref{tab:4}. The errors show that the proposed IGSP method performs well in aligning both the tiny model and the real-world datasets, with average rotation error less than 0.1 degree and translation error less than 0.1 meter, which provides a good foundation for further applications like 3D reconstruction and object extraction.

\begin{figure}
\begin{center}
\includegraphics[width=1.0\linewidth]{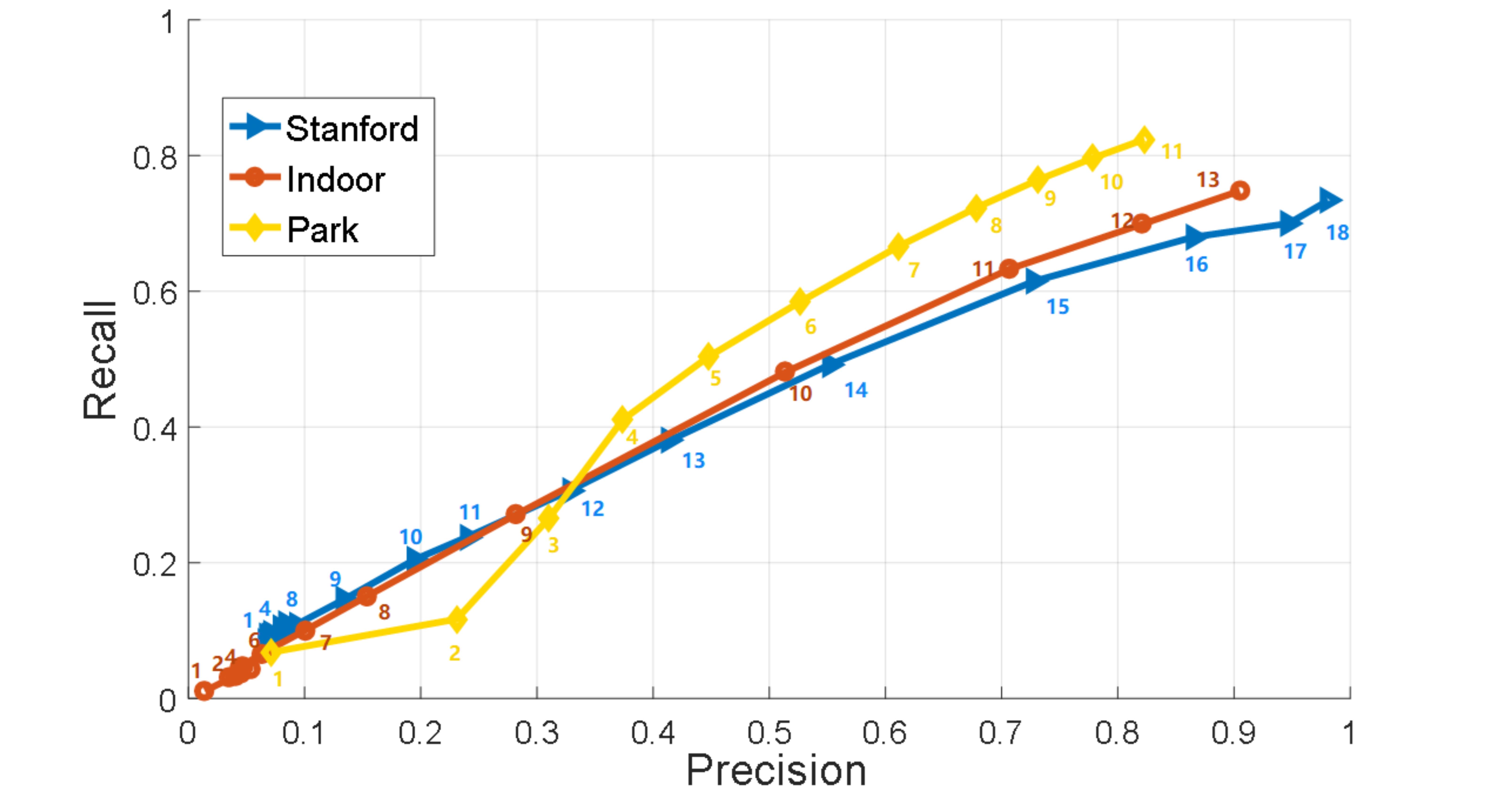}
\end{center}
   \caption{PR-Curve during iterative process on different Datasets}
\label{fig:8}
\end{figure}

{ \bf Accuracy analysis.}
Fig.\ref{fig:8} shows the precision-recall curve during IGSP's iterative process for three datasets, in which each point represents a temporal correspondence result. Since the global optimal correspondences are applied in each iteration, both the precision and recall of the correspondence increase through the process and finally exceed 0.75 on all these datasets, thus resulting in excellent registration performance.

{ \bf Time performance analysis.}
A time efficiency evaluation of the proposed IGSP method is  conducted with regard to the runtime in each step after proper keypoint detection parameters and Iterative convergence condition are set. Table \ref{tab:3} lists the the average iteration number $k$, time for registration preparation $T_1$, time for each iteration $T_2$ and the total runtime ($T=T_1+kT_2$). As analyzed before, the time complexity of KM algorithm is $O\left( n^3 \right)$, in which $ n $ is the key point number.

\begin{table}
\caption{Time performance of the proposed IGSP method.}
\begin{center}
\begin{tabular}{|c|c|c|c|c|}
\hline
Dataset&$T_{1}(s)$&$T_{2}(s)$&\#Iteration&$T(s)$\\
\hline
Stanford &3.31&0.53&15&11.26 \\
\hline
Park &8.12&5.65&13&81.57 \\
\hline
Indoor &6.25&3.92&14 &61.13\\
\hline
\end{tabular}
\end{center}
\label{tab:3}
\end{table}

{ \bf Performance comparison and analysis.}
\begin{table}
\caption{Registration accuracy and time performance comparison}
\begin{center}
\begin{tabular}{|l|l|l|l|l|}
\hline
Dataset&Method&$e^r$(mdeg)&$e^t$(mm)&T(s)\\
\hline
&ICP\cite{62}&/&/&5.6 \\
Stanford&3DNDT\cite{77}&/&/&\textbf{5.3} \\
30\% &Super4PCS\cite{Super4PCS}&74.21 &0.17 &10.5\\
overlapped&FM+GC\cite{DONG201861}&758.23&2.59&8.9\\
&IGSP&\textbf{9.89}&\textbf{0.02}&11.3\\
\hline
&ICP&/&/&9.4 \\
Park&3DNDT&/&/&{\textbf{8.1}} \\
65\%&Super4PCS&205.14&187.65&38.0\\
overlapped&FM+GC& 184.56& 414.61&35.2\\
&IGSP&{\textbf{93.74}}&\textbf{{85.12}}&81.6\\
\hline
&ICP&/&/&6.2\\
Indoor&3DNDT&/&/&\textbf{{6.0}} \\
70\%&Super4PCS&209.85 &/&47.1\\
overlapped&FM+GC&486.15 & 431.90&29.8\\
&IGSP&{\textbf{100.42}}&{\textbf{25.53}}&61.1\\
\hline
\end{tabular}
\end{center}
\label{tab:4}
\end{table}
To further analyze the performance of the proposed IGSP method, several pairwise point cloud registration methods (ICP\cite{62},3D-NDT\cite{77},Super4PCS\cite{Super4PCS} and feature matching with geometric consistency (FM+GC)\cite{DONG201861}) are selected for performance comparison using the Park dataset and the Indoor dataset. Key parameters of all the compared methods are set according to the parameter settings recommended in the original articles. 

Table \ref{tab:4} lists the average registration errors and runtime of the compared methods on three datasets, in which '$/$' means the registration failed (the value is great than 1000).  It is found that ICP and 3D-NDT fail when good initial alignment or prior knowledge is not provided. Super4PCS has poor performance on point clouds with limited overlapping and too many similar structures, especially for the indoor corridor. IGSP outperforms the FM+GC since many correct matching keypoint pairs are also rejected by geometric consistency filter, which leads to a relatively low recall of keypoints matching. However, the time efficiency of IGSP is inferior to all the compared methods due to the iteration process and the $O\left( n^3 \right)$ time complexity of KM algorithm, which will be the main concern of our future work.

\section{Conclusion and future work}
Nowadays, point cloud registration is the basis of many applications. This paper presented the Iterative Global Similarity Points (IGSP) algorithm, which iteratively find the corresponding keypoints considering global similarity and estimate the rigid body transformation to achieve coarse-to-fine pairwise point clouds registration. We validated its performance on different scenarios. Comprehensive experiments indicated that the proposed IGSP algorithm obtained good performance in correspondence precision, recall and registration accuracy. Although the proposed method provides satisfactory registration results, it is time consuming due to the iteration process and the $O\left( n^3 \right)$ time complexity of KM algorithm. In future work, we will try other efficient and reasonable methods (\eg. Minimum Cost Max Flow (MCMF) and Graph Cut) to solve the proposed energy function. A smooth term for geometric consistency in energy function will also be considered.


{\small
\bibliographystyle{ieee}
\bibliography{IGSP}
}

\end{document}